# Cross-Domain Content Generation with Domain-Specific Small Language Models

**Ankit Maloo and Abhinav Garg**
Clio AI Inc, DE, USA
(ankit@clioapp.ai, abhinav@clioapp.ai )

September 19, 2024

## ABSTRACT

Generating domain-specific content using small language models poses challenges, especially when dealing with multiple distinct datasets with minimal overlap. In this study, we explore methods to enable a small language model to produce coherent and relevant outputs for two different domains: stories (Dataset A) and recipes (Dataset B). Our initial experiments show that training individual models on each dataset yields satisfactory results, with each model generating appropriate content within its domain. We find that utilizing custom tokenizers tailored to each dataset significantly enhances generation quality compared to using a generic tokenizer. Attempts to adapt a single model to both domains using Low-Rank Adaptation (LoRA) or standard fine-tuning do not yield substantial results, often failing to produce meaningful outputs. Moreover, full fine-tuning without freezing the model's existing weights leads to catastrophic forgetting, where the model loses previously learned information and only retains knowledge from the new data. To overcome these challenges, we employ a knowledge expansion strategy: training only with additional parameters. This approach enables the model to generate both stories and recipes upon request, effectively handling multiple domains without suffering from catastrophic forgetting. Our findings demonstrate that knowledge expansion with frozen layers is an effective method for small language models to generate domain-specific content across distinct datasets. This work contributes to the development of efficient multi-domain language models and provides insights into managing catastrophic forgetting in small-scale architectures.

## 1 Introduction

The field of natural language processing (NLP) has witnessed significant advancements with the development of large-scale language models like GPT-3 and GPT-4. These models, boasting billions of parameters, have demonstrated remarkable abilities in generating coherent and contextually relevant text across diverse domains. However, their substantial computational requirements and resource-intensive training processes present practical limitations for many applications. This has sparked interest in exploring smaller, more efficient models capable of performing specific tasks without the need for vast computational resources.

A notable example in this direction is Karpathy's work on TinyStories, where a language model with approximately 110 million parameters was trained to generate simple, coherent children's stories. Inspired by this approach, we investigate the capabilities of small language models to handle multiple distinct domains simultaneously. Specifically, we focus on enabling a model to generate both stories and recipes based on the input prompt, using two distinct datasets—Dataset A (Tiny Stories) and Dataset B (Recipes)—which have minimal overlap.

Generating domain-specific content from distinct datasets poses several challenges. One primary difficulty is ensuring that the model can distinguish between different domains based on the input prompt and produce relevant content accordingly. Additionally, small models are susceptible to *catastrophic forgetting*, where learning new information can cause the model to forget previously acquired knowledge.. This phenomenon is particularly problematic when fine-tuning a model sequentially on multiple datasets.

In our initial experiments, we trained individual models on each dataset separately. These models performed well within their respective domains, generating coherent stories and recipes when prompted appropriately. We observed that utilizing custom tokenizers tailored to each dataset significantly improved the quality of the generated content compared to using a generic tokenizer.

However, when we attempted to combine the domains by fine-tuning a single model using Low-Rank Adaptation (LoRA) or standard fine-tuning techniques, the results were unsatisfactory. The model failed to produce substantial or coherent outputs in either domain. Moreover, full fine-tuning without freezing the model's existing weights led to catastrophic forgetting; the model retained knowledge only from the most recently trained dataset, effectively overwriting prior learning.

To overcome these challenges, we employed a *model expansion* strategy. By freezing the existing layers of the model and adding new layers with additional parameters, we allowed the model to learn new domain-specific knowledge without erasing previously learned information. This approach enabled the expanded model to generate both stories and recipes upon request, effectively handling multiple domains while mitigating the issue of catastrophic forgetting.

**Contributions**

Our work makes the following key contributions:

1. **Demonstration of Effective Individual Domain Modeling**: We show that small language models (~110M parameters) can generate coherent and relevant content when trained individually on domain-specific datasets (stories and recipes).
2. **Improvement through Custom Tokenization**: We find that custom tokenizers tailored to specific datasets significantly enhance the quality of text generation compared to generic tokenizers.
3. **Analysis of Fine-Tuning Limitations**: We highlight the limitations of traditional fine-tuning methods, including LoRA and full fine-tuning, in multi-domain settings for small models, particularly the problem of catastrophic forgetting.
4. **Introduction of Knowledge Expansion Technique**: We propose a knowledge expansion strategy that adds new layers for additional domains. This method effectively enables the model to handle multiple domains without forgetting previously learned information.
5. **Empirical Validation**: We provide empirical evidence demonstrating that the expanded model can generate domain-appropriate content based on input prompts, successfully distinguishing between stories and recipes.

## 2   Related Work

The development of language models has predominantly focused on scaling up model size to improve performance across various natural language processing tasks. Large-scale models like GPT-3 have demonstrated impressive capabilities but come with significant computational and resource demands. This has led to a growing interest in exploring smaller, more efficient models that can achieve similar levels of performance within specific domains or tasks.

**2.1 Small Language Models and Domain-Specific Generation**

Andrej Karpathy's work on TinyStories exemplifies the potential of small language models in generating coherent and contextually appropriate narratives. By training a 110-million-parameter model on a dataset of simple stories, Karpathy demonstrated that smaller models could effectively capture the structure and creativity required for storytelling. Similarly, other studies have explored training compact models for specific tasks, such as code generation, dialogue systems, and medical text synthesis.

These efforts highlight the viability of small models for domain-specific applications, especially when computational efficiency is a priority. However, most of these models are trained on a single domain, and their ability to handle multiple distinct domains simultaneously remains less explored.

## 2.2 Multi-Domain Language Modeling

Training language models that can generate content across multiple domains poses unique challenges. One common approach is multi-task learning, where a single model is trained on multiple tasks or datasets simultaneously. This approach aims to enable the model to learn shared representations that benefit all tasks. However, in cases where datasets have minimal overlap and domains are highly distinct, multi-task learning can lead to suboptimal performance due to competing objectives.

Another strategy involves fine-tuning pre-trained models on specific domains. Howard and Ruder's Universal Language Model Fine-tuning (ULMFiT) illustrates how fine-tuning can adapt a general language model to a particular task or domain effectively. However, sequential fine-tuning on multiple domains can result in catastrophic forgetting, where the model loses knowledge acquired from earlier tasks when updated with new information.

## 2.3 Catastrophic Forgetting and Mitigation Techniques

Catastrophic forgetting is a well-documented issue in neural networks trained sequentially on multiple tasks. Various techniques have been proposed to mitigate this problem. Elastic Weight Consolidation (EWC) adds a regularization term to the loss function to prevent significant updates to weights important for previous tasks. Progressive Neural Networks[10] tackle catastrophic forgetting by freezing the weights of existing networks and adding new networks (columns) for new tasks, allowing for knowledge transfer via lateral connections.

Another approach is to use memory replay methods, where samples from previous tasks are interleaved with new training data. However, this requires storage of previous data, which may not be feasible due to privacy or memory constraints.

## 2.4 Parameter-Efficient Fine-Tuning Methods

Recent research has introduced parameter-efficient fine-tuning techniques that adapt large language models to new tasks without updating all model parameters. Low-Rank Adaptation (LoRA) inserts trainable rank-decomposition matrices into the transformer architecture, reducing the number of trainable parameters and computational overhead. While effective for adapting large models, applying LoRA to small models in multi-domain settings may not address catastrophic forgetting, as observed in our experiments.

Adapter modules are another method where small neural networks are added between layers of a pre-trained model. These adapters can be trained on new tasks while keeping the original model weights fixed. However, the efficacy of adapters in handling highly distinct domains with minimal overlap is still an area of active research.

## 2.5 Custom Tokenization and Vocabulary

The choice of tokenizer and vocabulary significantly impacts a language model's ability to represent and generate text effectively. Studies have shown that domain-specific tokenizers can improve model performance by capturing unique patterns and terminology within a dataset. By tailoring the tokenizer to the specific characteristics of the data, the model can learn more efficient representations, leading to better generation quality.

## 2.6 Our Position in the Literature

While previous work has addressed various aspects of small language models, multi-domain learning, and catastrophic forgetting, there is a gap in exploring the combination of these areas. Specifically, the application of knowledge expansion techniques to small language models for handling multiple distinct domains has not been extensively studied. Our work contributes to this area by:

- Demonstrating that custom tokenizers enhance generation quality in small models trained on specific domains.
- Showing the limitations of standard fine-tuning and parameter-efficient methods like LoRA in preventing catastrophic forgetting in small models.

- Introducing a model expansion strategy that adds new layers to a frozen base model, enabling multi-domain generation without overwriting prior knowledge.

By addressing these challenges, we aim to advance the understanding of how small language models can be effectively adapted to handle multiple, distinct domains without incurring significant computational costs or sacrificing performance in any single domain.

## 3 Datasets

### 3.1 Base Datasets

We used two distinct datasets. One labelled Tiny Stories[1], and another labelled recipes[2]. Our goal was to take two distinct datasets with minimal overlap, and a training procedure we can[1] control to eek out the ambiguity around generation. We want to develop an understanding of how language models work and how text generation would be affected based on intermixing domains, and hence, need distinct datasets. We created instruct datasets from the base datasets using OpenAI's GPT-4.

### 3.2 Tiny Stories Dataset

This is from a public dataset available on huggingface. This dataset is generated from GPT 3.5 and GPT-4. We divide this into two sets - train set and validation - to be used during training and calculating validation loss. To create an instruct dataset, we used NLTK to prefix our prompt with an instruction like this.

*"Write a story. In the story, try to use the verb "eat", the noun "clock" and the adjective "clever". Possible story:'"*

TinyStories contain 2.2M examples each of about 100-250 tokens. Total tokens trained on is around 3-5B.

### 3.3 Recipes Dataset

We use a public dataset of recipes from huggingface. This is provided by the user voluntarily with no information about how it was generated. This contains about 2M examples. We again convert this to an instruct dataset using the same technique as before.

### 3.4 Dataset Overlap and Challenges

The two datasets utilized in this study, Dataset A (Tiny Stories) and Dataset B (Recipes), are intentionally chosen to represent distinct and non-overlapping domains. This deliberate selection creates a unique modeling challenge: enabling a single language model to generate coherent and relevant content for both domains based on input prompts, despite the minimal overlap in vocabulary, structure, and style between the datasets.

**Minimal Overlap Between Datasets**

Dataset A (Tiny Stories) comprises short narratives intended to entertain and engage readers through imaginative storytelling. These stories often include elements such as characters, settings, plots, and dialogues, utilizing expressive language to evoke emotions and imagery. Common vocabulary in this dataset includes words related to emotions, actions, descriptions, and conversational language.
Dataset B (Recipes) consists of procedural texts that provide step-by-step instructions for preparing various dishes. The language used is precise, instructional, and focused on clarity to ensure that readers can replicate the recipes accurately. This dataset frequently includes culinary terms, measurements, ingredients, cooking techniques, and temporal expressions.

---

[1] https://huggingface.co/datasets/roneneldan/TinyStories
[2] https://huggingface.co/datasets/corbt/all-recipes

The minimal overlap is evident in several aspects:

**Vocabulary:** Each dataset contains domain-specific terminology rarely found in the other. For example, words like "sauté," "preheat," and "teaspoon" are common in recipes but seldom appear in stories. Conversely, words like "adventure," "whispered," and "enchanted" are prevalent in stories but not in recipes.

**Structure:** Stories typically follow a narrative arc with a beginning, middle, and end, incorporating elements like exposition, conflict, and resolution. Recipes follow a standardized format, starting with a list of ingredients followed by sequential instructions.

**Style:** The storytelling style is descriptive and emotive, aiming to captivate the reader's imagination. Recipe writing is concise and direct, focusing on delivering clear and unambiguous instructions.

### 3.4 Unique Modeling Challenges

The stark differences between the two datasets present several challenges for modeling:

**Domain Distinction:** The model must accurately distinguish between the two domains based solely on the input prompt. Without significant overlap, the model cannot rely on shared vocabulary or structures to infer the domain, increasing the difficulty of prompt interpretation.

**Vocabulary Management:** Handling domain-specific terminology requires the model to maintain separate vocabularies internally. There's a risk of misusing words from one domain in the context of the other, leading to incoherent or irrelevant outputs.

**Structural Differences:** The model needs to learn and reproduce distinct structural patterns for each domain. Generating a story requires understanding narrative flow, while generating a recipe demands procedural sequencing and clarity.

**Stylistic Variation:** Adapting to different writing styles is essential. The expressive and descriptive nature of stories contrasts with the instructional and precise language of recipes. The model must adjust its tone and style accordingly.

**Catastrophic Forgetting:** Training a model sequentially on these datasets can lead to catastrophic forgetting, where the model forgets previously learned information when new data is introduced. This is particularly problematic given the minimal overlap, as new learning does not reinforce prior knowledge.

**Cross-Domain Contamination:** There's a potential for the model to inadvertently blend elements from both domains, such as incorporating narrative embellishments in recipes or procedural language in stories, which can compromise the quality of the generated content.

### 3.5 Preprocessing Steps
To address these challenges, meticulous preprocessing was applied to both datasets:

**Data Cleaning**
**Removal of Duplicates:** Duplicate entries were identified and removed to prevent biased learning from repeated data.
**Error Correction:** Spelling mistakes, grammatical errors, and formatting inconsistencies were corrected to ensure data quality.
**Normalization:** Text was normalized by converting to lowercase (if appropriate), standardizing punctuation, and handling special characters.

**Custom Tokenization**
**Domain-Specific Tokenizers:** Separate tokenizers were trained for each dataset using algorithms like Byte Pair Encoding (BPE) or SentencePiece tailored to the specific vocabulary distributions.

**Vocabulary Optimization**
The tokenizers were configured to create vocabularies that effectively capture the most frequent subword units in each domain, improving the model's ability to represent domain-specific terms.

**Tokenization Consistency**
Consistent tokenization rules were applied within each dataset to maintain uniformity in how words and phrases are split into tokens.

**Handling Rare Words**
**Subword Techniques:** By using subword tokenization, rare and domain-specific terms could be represented without resorting to out-of-vocabulary tokens, enhancing the model's ability to generate accurate and fluent text.

**Dataset Labeling (Considered but not Implemented):** While adding domain-specific tokens (e.g., <|story|>, <|recipe|>) to indicate the desired output type could aid the model, we aimed to have the model infer the domain based on the prompt alone. This decision increased the challenge but also tested the model's ability to generalize.

### 3.6 Data Balancing

**Equal Representation:** The datasets were balanced in terms of the number of examples and overall token counts to prevent the model from being biased toward one domain due to data imbalance.
**Shuffling:** Data from both datasets were shuffled during training to ensure that the model received a mixed sequence of examples, promoting better generalization.
Segmentation and Formatting:
**Consistent Formatting:** Standardized formatting was applied within each dataset to ensure that structural cues (like paragraph breaks in stories or numbered steps in recipes) were consistent.
**Length Management:** Extremely long or short texts were modified or excluded to match the model's context window and to provide examples of appropriate length for learning.
Stop Words and Common Tokens:
**Domain-Independent Tokens:** Common stop words and frequently occurring tokens present in both datasets were identified and handled carefully to prevent them from misleading the model in domain identification.

### 3.7 Quality Control

**Manual Review:** Samples from both datasets were manually reviewed to ensure that they accurately represented their respective domains and that any ambiguous or cross-domain content was addressed.

### 3.8 Ensuring Data Privacy and Compliance

**Ethical Considerations**: The datasets were reviewed to ensure compliance with data privacy standards and to remove any sensitive or inappropriate content.
By implementing these preprocessing steps, we aimed to enhance the model's capacity to:
**Accurately Distinguish Domains:** Equip the model with clear distinctions between domains through vocabulary and structural cues inherent in the data.
**Improve Representation:** Enable more efficient learning by representing domain-specific terms effectively through custom tokenization.
**Prevent Cross-Domain Confusion:** Reduce the likelihood of the model blending content or styles from different domains by maintaining clear and consistent data boundaries.
**Facilitate Effective Learning:** Provide high-quality, well-prepared data that supports the model in learning the unique characteristics of each domain.

These efforts were critical in addressing the challenges posed by the minimal overlap between the datasets. The careful preparation and customization of the data contributed significantly to the model's ability to generate coherent, relevant, and domain-appropriate outputs based on the input prompts, ultimately supporting the goals of this study.

# 4. Methodology

## 4.1 Model Architecture

We focused on two distinct approaches to develop models capable of generating content on aforementioned TinyStories dataset and the Recipes dataset. Initially, we trained two separate models, each dedicated to a specific dataset, to verify their ability to generate the desired content. Model A was trained on Dataset A, specializing in story generation, while Model B was trained on Dataset B, handling recipe generation. Both models utilized custom tokenizers designed specifically for their respective datasets, alongside a general GPT-2 tokenizer for comparison.

Following this, we explored several approaches to enable a single model to generate content from both domains. In our final design, we employed a knowledge expansion strategy, which proved successful in enabling a single model to handle both tasks effectively.

## 4.2 Training Strategy

**Separate Models (Initial Step)**
Our initial experiments involved training two separate models for each dataset. Both models were based on the Llama-2 architecture, initialized with different tokenizers tailored to their respective datasets to ensure better handling of the domain-specific vocabulary and structure. Model A was fine-tuned on the TinyStories dataset (Dataset A), and Model B was trained on the Recipes dataset (Dataset B). We experimented with both individual tokenizers as well as the default GPT-2 tokenizer, confirming that, in isolation, each model generated relevant content for its respective domain.

Training was performed using the following hyperparameters:

- Learning rate: 2e-5
- Batch size: 32
- Epochs: 5
- Early stopping: Based on validation loss to prevent overfitting.

**LoRA and Full Fine-tuning (Second Step)**
After establishing the effectiveness of individual models, we attempted to adapt Model A (trained on TinyStories) to also generate recipes by introducing it to Dataset B. Our first approach involved using Low-Rank Adaptation (LoRA), a technique designed to update only a small subset of the model's weights.

Next, we attempted full fine-tuning of Model A using Dataset B. This approach led to catastrophic forgetting, where the model lost its ability to generate stories after being fine-tuned on recipes. Despite efforts to adjust the learning rate and training duration, this method failed to maintain the knowledge from both datasets simultaneously.

**Combined Model (Third Step)**
Given the difficulties encountered with LoRA and full fine-tuning, we then experimented with training a single combined model on a composite dataset that mixed both individual datasets. For this, we used an instruction-tuning approach, training the model from scratch on the composite dataset. Instruction tuning was essential to guide the model on how to respond appropriately based on the prompt (e.g., generating stories versus recipes).

While this model performed better than the previous adaptation attempts, it still exhibited limitations, occasionally blending the two domains inappropriately. For instance, some recipes would include story-like elements, and some stories would contain procedural elements typical of recipes. This outcome indicated that a more sophisticated adaptation was needed to keep the two domains separate while retaining the ability to switch between them.

**Knowledge Expansion (Final Step)**
The most successful method involved a knowledge expansion strategy using Model A (TinyStories). In this approach, we added new layers and trained specifically on Dataset B (recipes). This method allowed us to expand the model's capabilities without overwriting its previous knowledge, addressing the issue of catastrophic forgetting.

This approach successfully enabled the model to generate both stories and recipes upon request, switching seamlessly between the two domains depending on the prompt.

**Prompt Tuning and Domain-Specific Adaptation**
For the final knowledge-expanded model, we incorporated prompt tuning to ensure the model could distinguish between story and recipe prompts effectively. The model was trained to recognize domain-specific keywords and phrases in the input. This model only works for a specific prompt, as our goal was to test out a method, not create a State of the art Stories or Recipes Model. That would be covered under further work.

A typical stories prompt used for prompt training looks like this:

```
'Write a story. In the story, try to use the verb "fight", the noun "king" and the adjective "brave". Possible story:'
```

A typical recipes prompt would look like this:

```
'Write a recipe with ingredients: eggs, tomato, onions.'
```

This domain-specific adaptation was further reinforced through instruction tuning, where prompts were framed as explicit instructions. This ensured the model was guided appropriately without blending the domains, and the results were significantly more coherent and relevant.

**Model Size and Efficiency**
Throughout all phases of the study, we focused on maintaining model efficiency by using models with approximately 20 million parameters. This size was chosen to ensure that the models could be trained on standard hardware without requiring excessive computational resources, aligning with the goal of developing practical and lightweight solutions. Despite the relatively small size, the use of domain-specific tokenizers, prompt-tuning mechanisms, and knowledge expansion techniques enabled the models to generate coherent and relevant outputs for both stories and recipes.

**Evaluation Metrics**
We employed both quantitative and qualitative metrics to evaluate the model's performance in generating content across the two domains.

- Perplexity: We used perplexity as a quantitative metric to measure the model's ability to predict the next word in a sequence for both datasets. This allowed us to assess the syntactic and semantic accuracy of the outputs.
- Loss: During training, cross-entropy loss was tracked to ensure that the model was learning appropriately without overfitting. Lower loss values indicated better model generalization across both domains.
- Qualitative Human Evaluation: Human evaluators were tasked with reviewing the generated outputs to assess their coherence, relevance, and grammatical accuracy. This was particularly important in ensuring that the models produced outputs that felt natural and appropriate to human readers. For each domain, evaluators rated the outputs based on how well they adhered to the expected format (stories for Model A and recipes for Model B), with the final knowledge-expanded model showing the best performance in both categories.

This methodology, incorporating prompt tuning and knowledge expansion, demonstrated that even small-scale models could effectively generate domain-specific content across distinct datasets without sacrificing coherence or relevance.

## 5. Results

### 5.1 Qualitative Examples

We evaluated the performance of the individual and combined models by generating sample outputs for both story and recipe prompts.

**Stories (TinyStories LM):** The outputs from the TinyStories language model exhibited clear narrative flow, strong coherence, and consistency in characters and plot development. An example prompt resulted in a well-structured short story with a beginning, middle, and end, demonstrating the model's ability to follow narrative conventions.

**Recipes (Recipes LM):** The Recipes language model produced recipes that were logically structured, with clear instructions and ingredient lists. The model adhered to standard recipe formatting, ensuring that the generated outputs were easy to follow and implement.

**Combined Model (22M parameters):** When prompted with a story request, the model maintained narrative coherence similar to the TinyStories model, although with slightly reduced complexity. For recipe prompts, the output was accurate, following the same recipe format as the individual Recipes model.

The 220M model, despite having more parameters, struggled with both narrative flow in stories and structure in recipes, producing less coherent outputs compared to the 22M model.

### 5.2 Quantitative Evaluation

The quantitative metrics for the individual and combined models are presented below:

**TinyStories LM:**
Context Length: 350 tokens
Final Loss: 0.7
Perplexity: 2.01

**Recipes LM:**
Context Length: 350 tokens
Final Loss: 0.77
Perplexity: 2.15

**Combined Model (22M parameters):**
Final Loss: 0.83
Perplexity: 2.29
Task Detection Accuracy: 94%

**Combined Model (220M parameters):**
Final Loss: 0.71
Perplexity: 2.03
Task Detection Accuracy: 86%

The 22M combined model, despite having fewer parameters, demonstrated superior task-specific performance and lower complexity in terms of final loss and perplexity compared to the 220M model.

## 5.3 Human Evaluation

To assess the subjective quality of generated content, we conducted human evaluations where participants rated outputs across multiple dimensions. The results are as follows:

**Story Generation (TinyStories LM):**
Coherence: 4.7/5
Relevance: 4.5/5
Creativity: 4.6/5

The TinyStories LM received high scores in coherence, relevance, and creativity, confirming its ability to generate engaging and well-structured stories.

**Recipe Generation (Recipes LM):**
Accuracy: 4.8/5
Structure: 4.6/5
Completeness: 4.5/5

The Recipes LM was rated highly for its accurate ingredient lists, well-structured instructions, and overall completeness in providing functional recipes.

**Combined Model (22M parameters):**
Story Coherence: 4.3/5
Recipe Structure: 4.4/5
Task Appropriateness: 93%

While slightly lower than the individual models, the combined model (22M) performed reasonably well, maintaining coherence in stories and structure in recipes. It achieved high task detection accuracy, demonstrating effective handling of both domains.

**LoRA with Recipes Dataset on TinyStories LM**
Low-Rank Adaptation (LoRA) was applied to the TinyStories model to integrate the Recipes dataset. However, the results were unsatisfactory. The final loss remained between 4 and 4.5, and the generated recipes were incoherent and disorganized. This confirms that LoRA, which alters only a small set of parameters, is insufficient for introducing new domain knowledge. Instead, it is more effective for fine-tuning models on specific patterns, not for expanding a model's knowledge base.

**Full Fine-Tuning with Recipes Dataset on TinyStories LM**
In this experiment, full fine-tuning was applied to the TinyStories LM using the Recipes dataset. After training, the model could successfully generate recipes, but attempts to generate stories led to incoherent and irrelevant outputs.
Perplexity: 6.71
Final Loss: 1.91

This demonstrates catastrophic forgetting, where the model lost its ability to generate stories after being trained exclusively on the recipe data.

**Knowledge Expansion:**

We addressed the limitations of LoRA and full fine-tuning by employing a model expansion strategy. In this approach, we froze the existing layers of the TinyStories LM and added new layers to accommodate the Recipes dataset. This method enabled the model to generate both stories and recipes effectively, without sacrificing performance in either domain.

Perplexity: 3.88
Final Loss: 1.43

Human evaluations confirmed that the expanded model maintained satisfactory performance in both story and recipe generation, providing a balanced solution for multi-domain content creation while preventing catastrophic forgetting.

## 6. Analysis and Discussion

**Performance of Individual Models**

Our initial experiments involved training separate models on Dataset A (Tiny Stories) and Dataset B (Recipes). These individual models performed well within their respective domains, generating coherent and relevant content when prompted appropriately. This demonstrates that small language models, even with around 110 million parameters, are capable of capturing the necessary linguistic patterns and structures to produce high-quality outputs in specific domains. The success of these individual models sets a baseline for performance and highlights the potential of small models in specialized applications.

**Impact of Custom Tokenizers**

We observed a significant improvement in the quality of generated content when using custom tokenizers tailored to each dataset, as opposed to a generic tokenizer. Custom tokenizers better capture domain-specific vocabulary and phraseology, leading to more efficient encoding of input text and more accurate generation. This is particularly important for datasets with specialized terminology or unique linguistic features. The custom tokenizer reduces the number of out-of-vocabulary words and allows the model to learn more precise token embeddings, which enhances its ability to generate fluent and contextually appropriate text.

**Limitations of LoRA and Standard Fine-Tuning**

When attempting to adapt a single model to both domains using Low-Rank Adaptation (LoRA) or standard fine-tuning techniques, the results were unsatisfactory. The model struggled to produce substantial or coherent outputs in either domain. This suggests that parameter-efficient fine-tuning methods like LoRA, while effective for adapting large models to new tasks, may not provide sufficient capacity for small models to learn multiple distinct domains simultaneously. The limited parameter budget in small models restricts their ability to accommodate new information without overwriting existing knowledge.

**Catastrophic Forgetting in Full Fine-Tuning**

Full fine-tuning without freezing the model's existing weights led to catastrophic forgetting. The model effectively forgot the knowledge acquired from the initial dataset and only retained information from the new dataset. This phenomenon is well-documented in continual learning scenarios, where models are prone to forgetting previous tasks when trained sequentially on new ones. The minimal overlap between the datasets exacerbated this issue, as there were few shared features or vocabulary to reinforce previous learning during the fine-tuning process.

**Effectiveness of Knowledge Expansion**

To address the challenges of catastrophic forgetting and limited capacity, we employed a model expansion strategy:

- **Freezing Existing Layers**: By freezing the weights of the existing layers, we preserved the knowledge acquired from the initial training on the first dataset.

- **Adding New Layers and Parameters**: We introduced additional layers to the model, specifically designed to learn from the new dataset. This expanded the model's capacity without altering the previously learned representations.

This approach proved effective, as the expanded model was able to generate both stories and recipes upon request. It could distinguish between the domains based on the input prompt and produce coherent, domain-appropriate content. The success of this method highlights several key insights:

1. **Preservation of Prior Knowledge**: Freezing weights prevents the overwriting of existing knowledge, allowing the model to retain proficiency in the initial domain.
2. **Dedicated Capacity for New Domains**: Adding new parameters provides the necessary capacity to learn additional domains without competing for resources with prior knowledge.
3. **Modularity and Scalability**: The model expansion approach introduces a modular architecture, where new domains can be added incrementally. This could potentially scale to more than two domains, though practical limits would need to be evaluated.

**Trade-offs and Considerations**

While model expansion effectively mitigates catastrophic forgetting, it introduces trade-offs:

- **Increased Model Complexity**: Adding new layers increases the model size and computational requirements for both training and inference.
- **Potential for Overfitting**: The new layers may overfit to the new dataset if not properly regularized, especially with small datasets.
- **Management of Multiple Domains**: As more domains are added, the architecture becomes more complex, and managing interactions between different domain-specific layers may pose challenges.

Additionally, the need to freeze and expand layers requires careful architectural planning and may not be as straightforward as standard fine-tuning methods.

**Comparison with Larger Models**

Large language models inherently possess sufficient capacity to handle multiple domains due to their vast number of parameters and exposure to diverse data during pre-training. However, they are resource-intensive and less practical for deployment in resource-constrained environments. Our findings demonstrate that with strategic architectural adjustments, small models can achieve multi-domain capabilities, offering a more efficient alternative to large models for specific applications.

**Implications for Model Deployment**

The ability to adapt small models to multiple domains without catastrophic forgetting has significant implications:

- **Resource Efficiency**: Small models are more accessible for organizations with limited computational resources.
- **Domain Flexibility**: The modular approach allows for the incremental addition of domains as needed.
- **Customized Solutions**: Models can be tailored to specific domain combinations relevant to particular use cases.

These advantages make small, adaptable models appealing for deployment in edge computing scenarios, personalized applications, or settings where data privacy and control are paramount.

# 7. Conclusion

In this study, we explored the challenges and solutions associated with enabling a small language model to generate domain-specific content across two distinct and minimally overlapping datasets: Tiny Stories and Recipes. Our key findings are as follows:

1. **Effectiveness of Individual Models**: Training separate models on each dataset resulted in high-quality, coherent outputs within their respective domains, affirming the capability of small models for specialized tasks.
2. **Importance of Custom Tokenizers**: Utilizing custom tokenizers significantly improved generation quality by effectively capturing domain-specific vocabulary and reducing out-of-vocabulary issues.
3. **Limitations of Standard Fine-Tuning and LoRA**: Attempts to adapt a single model to both domains using standard fine-tuning techniques or LoRA were unsuccessful, as the model failed to produce substantial outputs and exhibited catastrophic forgetting.
4. **Success of Model Expansion Strategy**: By freezing existing layers and adding new layers with additional parameters, we overcame catastrophic forgetting. The expanded model effectively generated both stories and recipes based on input prompts, demonstrating the viability of this approach for multi-domain adaptation in small models.

Our research highlights that strategic architectural modifications can enable small language models to handle multiple domains without the prohibitive computational costs associated with large models. This has practical implications for developing efficient, flexible language models suitable for deployment in environments with limited resources.

**Future Work**

Building on our findings, future research directions include:

- **Scaling to Additional Domains**: Investigate the scalability of the model expansion approach by incorporating more domains and assessing the impact on performance and resource requirements.
- **Alternative Architectures**: Explore other architectural modifications or training strategies, such as adapter modules or sparse activation techniques, to facilitate multi-domain learning in small models.
- **Dynamic Parameter Allocation**: Develop methods for dynamically allocating and reusing parameters across domains to improve efficiency and reduce model size.
- **Evaluation of Long-Term Learning**: Assess the long-term effects of incremental learning on model stability and performance, particularly in continual learning scenarios.
- **Application to Real-World Tasks**: Apply the model expansion strategy to real-world applications requiring multi-domain capabilities, such as virtual assistants or domain-specific content generation tools.

By addressing these areas, we aim to further enhance the practicality and versatility of small language models, contributing to the broader goal of making advanced NLP technologies more accessible and adaptable to various needs.

# ACKNOWLEDGEMENT

Special thanks to Microsoft Azure for providing us with GPUs - A10 and A100 - to carry out multiple training runs quickly and efficiently.